# Use of the C4.5 machine learning algorithm to test a clinical guideline-based decision support system

Jean-Baptiste LAMY [a,1], Anis ELLINI [a], Vahid EBRAHIMINIA [a]
Jean-Daniel ZUCKER [a], Hector FALCOFF [b] and Alain VENOT [a]

[a] *Laboratoire d'Informatique Médicale et de Bio-informatique (LIM&BIO), UFR SMBH, Université Paris 13, Bobigny, France*

[b] *Société de Formation Thérapeutique du Généraliste (SFTG), Paris, France*

**Abstract.** Well-designed medical decision support system (DSS) have been shown to improve health care quality. However, before they can be used in real clinical situations, these systems must be extensively tested, to ensure that they conform to the clinical guidelines (CG) on which they are based. Existing methods cannot be used for the systematic testing of all possible test cases.

We describe here a new exhaustive dynamic verification method. In this method, the DSS is considered to be a black box, and the Quinlan C4.5 algorithm is used to build a decision tree from an exhaustive set of DSS input vectors and outputs. This method was successfully used for the testing of a medical DSS relating to chronic diseases: the ASTI critiquing module for type 2 diabetes.

## Introduction

Clinical Guidelines (CG) are useful for physicians [1]. However, guidelines printed on paper are difficult to use efficiently during medical consultation [2]. This has led to the development of decision support systems (DSS) based on the CG [3]. The ASTI project in France provides an example of such a system. This DSS aims to improve therapeutic care for patients with chronic diseases, by helping physicians to take into account the recommendations expressed in CG [4]. ASTI includes a critiquing module — a rule-based system composed of a knowledge base and an inference engine [5, 6]. This module is automatically activated when the physician writes a drug prescription; it compares the treatment proposed by the physician with that recommended by the CG, and issues an alert if they differ. ASTI has been applied to several chronic diseases, including type 2 diabetes.

[1]Corresponding Author: Jean-Baptiste Lamy, LIMBIO, UFR SMBH, Université Paris 13, 74 rue Marcel Cachin, 93017 Bobigny cedex, France; E-mail: jibalamy@free.fr.

Before their use in real clinical situations, DSS such as ASTI must be extensively tested to ensure their medical validity. Errors may be encountered at various levels: in the knowledge base, in the inference engine or in the system specifications. Few papers on the testing of medical DSS have been published, although knowledge-based system verification has been widely studied outside the medical domain. Preece [7, 8] distinguishes between two types of verification method. *Static methods* do not require the DSS to be run. They involve the inspection of the knowledge base by an expert, or checking for syntactic errors, logical errors (*e.g.* unsatisfiable conditions) or semantic errors (*e.g.* a male patient being pregnant) in the knowledge base [9]. These methods may identify errors, but cannot ensure the total absence of errors [7, 10].

*Dynamic methods* involve the running of the DSS with a test base. The test base may be written manually, or using automatic methods aiming to identify the "most relevant" test cases [10, 11]. The intervention of a human expert is required to determine whether the responses of the DSS are satisfactory. These methods therefore cannot be used for the systematic testing of all possible test cases, as there are generally too many such cases for manual review by an expert.

We aimed to check the conformity of the ASTI critiquing module to the CG used to design it — the French CG for type 2 diabetes [12]. We present a new dynamic verification method for ``rebuilding'' the knowledge contained in the CG from an exhaustive set of test cases, using machine learning techniques to construct a decision tree. We applied this method to the ASTI critiquing module for type 2 diabetes, and present the results of a comparison, by an expert, of the generated decision tree with the original CG. Finally, we discuss the potential value of such a method and possibilities of applying this method to other DSS.

## 1 Methods

We propose a general verification method with three steps: (1) generation of an exhaustive set of possible input vectors for the DSS, and running of the DSS to determine the output for each input vector, (2) extraction of knowledge from the set of (input vector, output result) pairs by applying learning or generalization algorithms, and (3) comparison, by an expert, of the knowledge extracted in step 2 with the original source of knowledge (here, the CG).

### *1.1 Generating Input Vectors and Outputs*

It is possible to generate an exhaustive (or almost exhaustive) set of input vectors by considering a set of variables expressing the various elements of input for the DSS, and generating all possible combinations of the variables' values. Continuous variables (*e.g.* glycosylated haemoglobin) are limited to a few values, corresponding, for example, to the threshold values expressed in the CG. Finally, the output associated with each input vector is obtained by running the DSS.

### *1.2 Building the Decision Tree*

A decision tree is built from the input vectors and the associated outputs, using C4.5 [13], a reference algorithm in machine learning. Pruning must be disabled, to ensure 0% error in the tree. Factorization rules are applied to reduce the size of the tree: (1) if all the children of a given node include the same element of a recommendation (*e.g.* a recommended drug treatment), this element of information can be included in the node and removed from its children, (2) if a variable can take several values leading to the same recommendations, the largest set of such values can be grouped together as “<other>”.

### *1.3 Comparing the Decision Tree with Clinical Guidelines*

Finally, the decision tree is compared with the CG by experts. These experts should be medical experts briefly trained in the evaluation method and the reading of the decision tree. The experts must check that the treatments recommended by the tree conform to the CG, and should check the CG to ensure that none of the recommendations included in the CG are missing from the tree.

## 2 Results

The general method presented above was applied to the ASTI critiquing module for type 2 diabetes.

### *2.1 Generating Input Vectors*

The inputs of the ASTI critiquing module are related to (a) the patient's clinical condition, (b) the patient's treatment history, (c) the new treatment proposed by the physician, and (d) efficiency and the tolerance of the current treatment. For type 2 diabetes, these inputs correspond to the variables shown in Table 1. Doses were taken into account, using the following rule: if the treatment proposed is the same as the current treatment, the dose is understood to be modified: reduced if the problem identified was poor tolerance, and increased if the problem identified was low efficiency. A few rules were used to eliminate unrealistic combinations. For example, if the current treatment is diet, there cannot be drug intolerance.

Combinations of the values given in Table 1 generated an almost exhaustive set of input vectors. Only cases in which the patient had more than one drug intolerance, and cases involving quadritherapies (which are never recommended by the CG) were excluded. This approach yielded 147,680 input vectors.

### *2.2 Building the Decision Tree*

The data were initially processed so that the list of valid treatments was used as the classifying variable, rather than the critiquing module’s output (*i.e.* conform, not optimal or non conform). The decision tree was then built using the C4.5 algorithm. Before the factorization rules described in the methods section were applied, the decision tree included 87 nodes. After the application of these rules, the final tree included only 60 nodes (see Figure 1).

For a patient with a BMI of 25 kg/m² and an HbA1c level of 7%, with a current treatment of diet plus metformin shown to be inefficient, the decision tree recommends a first-line treatment of diet + metformin (with a higher dose of metformin as the recommended treatment is the same as the current treatment, see section 3.1), diet + glinide, diet + sulfonamide, diet + metformin + glinide or diet + metformin + sulfonamide, and a second-line treatment of diet + glinide + alphaglucosidaseinhibitors (AGI), diet + metformin + AGI, diet + sulfonamide + AGI, diet + metformin + glinide, diet + metformin + sulfonamide or diet + metformin + glitazone.

| **Variable** | **Definition** | **Retained values** |
|---|---|---|
| Diabetes discovery | Was diabetes discovered at an early or late stage ? | Early, late |
| BMI | Body mass index | <= 27 kg/m², > 27 kg/m² |
| HbA1c | Glycosylated hemoglobin levels | <= 6.5%, > 6.5% (*) |
| Current type of treatment | The type of the treatment currently administered | No treatment, diet only, monotherapy, bitherapy, tritherapy, single daily insulin treatment, fractioned insulin treatment |
| Current treatment | Treatment currently administered | 19 possible values, *e.g.*: no treatment, diet, diet+metformin |
| Problem | The medical problem preventing the continuation of current treatment | Low efficiency, poor tolerance |
| Efficiency | Level of efficiency of the current treatment (meaningful only if the problem is low efficiency) | Partial, null |
| Poorly tolerated drug | The drug in the current treatment responsible for intolerance (meaningful only if the problem is poor tolerance) | Metformin, sulfonamide, glinide, alpha-glucosidase inhibitors, glitazone, very slow-acting insulin, delayed-action insulin |
| Proposed treatment | The new treatment proposed by the physician, which may or may not conform to the CG | (As for current treatment) |

**Table 1:** The list of variables considered for type 2 diabetes, and the values retained. (*) Only two values were considered, as ASTI makes use of treatment efficiency to determine whether the treatment should be strengthened.

For a patient with a BMI of 29 kg/m² and an HbA1c level of 6.8%, with a current treatment of diet plus metformin, and who does not tolerate the metformin, the decision tree recommends a first-line treatment of diet + AGI, and a second-line treatment of diet + glinide or diet + sulfonamide. However, decreasing the dose of the metformin is not recommended.

### *2.3 Comparing the Decision Tree with Clinical Guidelines*

We asked two physicians to compare the decision tree with the original CG. The experts found the decision tree to be easily readable, although one of them was at first confused by the use of only two levels of treatment intention on the tree. The experts searched the tree for the therapeutic recommendations expressed in the CG, and found that none of them was missing. They also ensured that the various branches of the tree lead to the same prescription that the CG. Consequently, the experts considered the tree to conform to the CG.

```
+--problem = low efficiency: diet+metformin OR
|   |                       (diet+glinide+AGI OR
|   |                        diet+metformin+AGI OR
|   |                        diet+sulfonamide+AGI)
|   +--BMI <= 27: diet+glinide OR
|   |             diet+metformin+glinide OR
|   |             diet+metformin+sulfonamide OR
|   |             diet+sulfonamide OR
|   |            (diet+metformin+glitazone)
|   +--BMI >  27: diet+metformin+glitazone OR
|                (diet+glinide OR
|                 diet+metformin+glinide OR
|                 diet+metformin+sulfonamide OR
|                 diet+sulfonamide)
+--problem = poor tolerance:
    +--BMI <= 27: diet+glinide OR diet+sulfonamide
    |   +--poorly_tolerated_drug = metformin: diet+AGI
    |   +--poorly_tolerated_drug = <other>  : diet+metformin
    +--BMI >  27: (diet+glinide OR diet+sulfonamide)
        +--poorly_tolerated_drug = metformin: diet+AGI
        +--poorly_tolerated_drug = <other>  : diet+metformin
```

**Figure 1:** Part of the decision tree generated, limited to patients currently treated by monotherapy and with HbA1c levels > 6.5%. Treatments in brackets are recommended as second-line treatments only. Treatments on non-terminal nodes apply to all the child nodes. AGI = alpha glucosidase inhibitors.

## 3 Discussion and conclusion

The verification method proposed made it possible to test the ASTI critiquing module for type 2 diabetes in an almost systematic manner. It also gives an overview of the DSS reasoning. The knowledge extracted from the DSS was represented using rules and decision lists as alternatives to decision trees. Rules were more verbose than the tree. The decision list was a little more concise, but it was easier to compare the tree with the textual CG, due to the tree-like sectioning of the CG. Other algorithms could be used, *e.g.* for controlling the tree's optimality. We tried to insert errors by deleting arbitrary rules in ASTI's knowledge base; these errors were clearly visible in the decision tree and were found by the experts.

The method proposed here has several advantages. First, it permits almost systematic testing. ASTI was preliminary tested with a hand-written test based on 796 input vectors. The method used here involved about 150,000 input vectors — three orders of magnitude greater than initially tested. Second, this method tests not only the DSS knowledge base, but also the inference engine. The decision tree obtained is not a simple conversion of the rule-based knowledge base into a decision tree. Indeed, the inference engine of the ASTI critiquing module includes several hard-coded generic rules: *e.g.* if the patient tolerated a drug poorly in the past, it should not be prescribed again, unless its dose is decreased. The decision tree took into account the knowledge base, but also these generic rules, including any possible bugs in the inference engine.

Finally, the method proposed considers the DSS as a black box, and thus makes no assumptions about its internal functioning. This method could therefore be applied to many other DSS, independently from the DSS reasoning method: rule-based system, hard-coded rules, decision tree, neural network, or any other algorithm. All DSS based on a human-readable source of knowledge, such as a CG, are eligible (by opposition to *e.g.* a DSS using the k-nearest neighbours algorithm), which includes most of medical knowledge based DSS. To apply our testing method, one should first identify the DSS inputs, generate an almost exhaustive set of input vectors, and run the DSS for each of them. Then, one should extract knowledge from the input vectors and output data, using C4.5 or another learning algorithm. Finally, one should ask a human expert to compare the extracted knowledge to the knowledge that was used to build the DSS.

Three problems might be encountered when applying this method. First, it can be difficult to generate an almost exhaustive set of the DSS's input vectors, in particular when the number of variables is very high, *e.g.* for hypertension CG, or when there are many continuous variables. However medical recommendations usually define threshold values that allow to discretize continuous variable easily, as we have done for diabetes type 2. Second, it might be impossible to generate in a reasonable time the outputs for all input vectors (DSS too slow or too many vectors). Third, learning algorithms might be unadapted to the knowledge used by the DSS. For example, rule conditions such as ``if the patient has two or more risk factors from a given list..." cannot be learnt by C4.5. To solve this problem, one can use more sophisticated learning algorithms, *e.g.* based on description logic, or derive input variables from the other variables, *e.g.* the number of risk factors could be added as an input variable.

In conclusion, the ASTI critiquing module for type 2 diabetes was almost exhaustively tested in this study, using an original black-box dynamic verification method. This method appears generic and therefore applicable to other medical DSS.